\documentclass[letterpaper, 10 pt, conference]{ieeeconf}

\IEEEoverridecommandlockouts

\usepackage{etextools}
\usepackage{amssymb}
\usepackage{bm}
\usepackage{float}
\usepackage{makeidx}
\usepackage{amsmath}
\usepackage{bbm}
\usepackage{epsfig}
\usepackage{epsf}
\usepackage{psfrag}
\usepackage{verbatim}
\let\labelindent\relax
\usepackage{enumitem}
\usepackage{color}
\usepackage{multirow}
\usepackage{tabularx}
\usepackage{booktabs}
\usepackage[tight,footnotesize]{subfigure}
\usepackage{array}
\usepackage{soul}
\usepackage{footnote}
\usepackage{cite}
\usepackage{dblfloatfix}

\usepackage{color, colortbl}
\usepackage[colorlinks,bookmarksnumbered,citecolor=orange,urlcolor=orange]{hyperref}
\usepackage{graphicx}
\graphicspath{{Figures}}
\DeclareGraphicsExtensions{.pdf,.png}
\usepackage{bigstrut}
\usepackage[english]{babel}

\usepackage{csquotes}

\usepackage[T1]{fontenc}
\usepackage{algorithm, setspace}
\usepackage{algpseudocode}
\usepackage{url}
\usepackage{multirow}
\usepackage{float}
\usepackage{xcolor}
\usepackage{hyperref}
 \hypersetup{
     colorlinks=true,
     linkcolor=orange,
     filecolor=orange,
     citecolor=orange,      
     urlcolor=orange,
     }

\newcommand{\p}[1]{\smallskip \noindent \textbf{{#1}.}}
\newcommand{\eq}[1]{Equation~(\ref{eq:#1})}
\newcommand{\fig}[1]{Figure~\ref{fig:#1}}
\usepackage{balance}

\DeclareMathOperator*{\argmax}{arg\,max}

\title{\LARGE

Personalizing Interfaces to Humans with User-Friendly Priors

}

\author{Benjamin A. Christie, Heramb Nemlekar, and Dylan P. Losey
\thanks{This work is supported in part by NSF Grant $\#2129201$.}
\thanks{The authors are members of the Collaborative Robotics Lab (\href{https://collab.me.vt.edu/}{Collab}), Dept. of Mechanical Engineering, Virginia Tech, Blacksburg, VA 24061.
\newline
{e-mail: \texttt{\{benc00, hnemlekar, losey\}@vt.edu}}}
}

\begin{document}
\maketitle

\begin{abstract}

Robots often need to convey information to human users.
For example, robots can leverage visual, auditory, and haptic interfaces to display their intent or express their internal state.
In some scenarios there are socially agreed upon conventions for what these signals mean: e.g., a red light indicates an autonomous car is slowing down.
But as robots develop new capabilities and seek to convey more complex data, the meaning behind their signals is not always mutually understood: one user might think a flashing light indicates the autonomous car is an aggressive driver, while another user might think the same signal means the autonomous car is defensive.
In this paper we enable robots to \textit{adapt} their interfaces to the current user so that the human's personalized interpretation is aligned with the robot's meaning.
We start with an information theoretic end-to-end approach, which automatically tunes the interface policy to optimize the correlation between human and robot.
But to ensure that this learning policy is intuitive --- and to accelerate how quickly the interface adapts to the human --- we recognize that humans have \textit{priors} over how interfaces should function.
For instance, humans expect interface signals to be proportional and convex.
Our approach biases the robot's interface towards these priors, resulting in signals that are adapted to the current user while still following social expectations.
Our simulations and user study results across $15$ participants suggest that these priors improve robot-to-human communication. See videos here: \url{https://youtu.be/Re3OLg57hp8}

\end{abstract}


\section{Introduction}
\label{sec:intro}

\begin{figure}
\centering
\includegraphics[width=0.8\columnwidth]{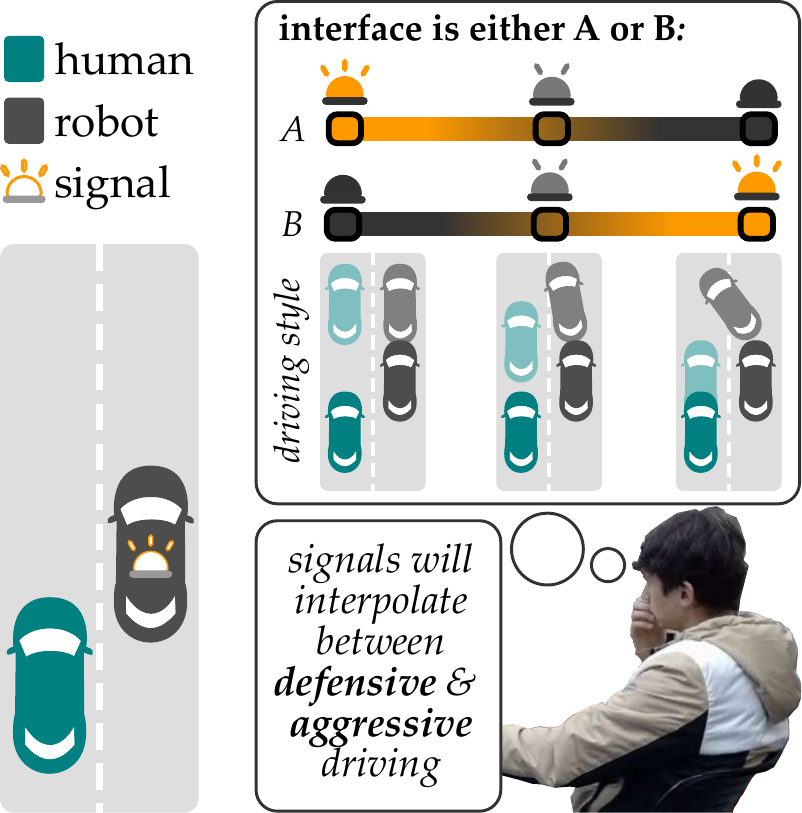}
    \caption{A human is attempting to pass an autonomous car. The autonomous car \textit{communicates} its policy (i.e., its driving style)
    using an LED mounted on the top of the car. Different users will interpret these LED signals in different ways. Although the interface does not know \textit{a priori} how any specific human will interpret its signals, we recognize that there are underlying patterns all users expect interfaces to follow.
    Here the interface signals should recognize that there are two intuitive options that interpolate between (at one extreme) defensive driving and (at the other extreme) aggressive driving.}
    \label{fig:front}
\vspace{-1.8em}
\end{figure}

Consider a robot communicating with a human.
For example, in \fig{front} a human driver is attempting to pass an autonomous car.
The human does not know the autonomous car's driving style --- is the autonomous car an aggressive driver (that will merge in front of the human) or a defensive driver (that will stay in its lane to give the human space)?
To communicate the robot uses an \textit{interface}.
In \fig{front} the autonomous car's interface is an LED light mounted on its roof that signals the robot's driving style.
This light can vary from fully on (an orange light) to fully off (a black light).
Importantly, there are multiple ways humans might interpret these signals: perhaps for one human an orange light means the robot is aggressive, while for another human the same signal indicates the robot is defensive.
To successfully communicate, here the robot must \textit{adapt} its interface signals to align with the current human's interpretation (e.g., switch between using orange or black for aggressive driving).

More generally, our work focuses on learning a mapping from the information the robot wants to convey to the signals the robot uses to convey that information.
In some scenarios this mapping is pre-established and mutually understood --- e.g., a red tail light communicates that a car is slowing down.
But we explore settings where humans interact with robotic interfaces (e.g., visual, auditory, haptic), and the meaning behind the interface's signals is not clear.
Existing works address this challenge in two ways.
On the one hand, some methods \cite{reddy2022first,kaupp2010human,dragan2013legibility, huang2019enabling,  brown2019machine,lee2021machine} assume the robot has a model of either the human or the task, and fine-tune the interface's signals based on that model.
But these methods require domain-specific knowledge and can result in mappings that are only able to convey a single task.
On the other hand, approaches like \cite{li2020learning} and \cite{christie2023limit} learn interface signals from scratch by gathering data from the operator across multiple interactions.
This removes the need to know the human's task --- but learning an interface end-to-end is time-consuming, and the learned mappings can turn out to be complex and unintuitive.

In this paper we bridge the gap between these approaches.
We seek to learn a task-invariant interface which can (a) communicate information without needing to know the human's intent while also (b) adapting rapidly and intuitively to the current user.
To achieve this goal, our hypothesis is that:
\begin{center}\vspace{-0.3em}
\textit{Although different humans may have different interpretations of the same signal, there are underlying patterns all users expect interfaces to follow.}
\vspace{-0.3em}
\end{center}
Consider our running example from \fig{front}.
We naturally expect one extreme of the LED light to indicate the robot is aggressive, the other extreme to indicate it is defense, and the intermediate signals should interpolate between aggressive and defensive driving.
In what follows, we encode this idea as \textit{priors} over the space of interface mappings.
We then derive an information theoretic approach that adapts signals to the current user while ensuring that the interface satisfies the given priors.
This leads to interfaces that are \textit{self-adjusting}: these interfaces tune their own signals to help the human correctly interpret the robot's information.
Returning to \fig{front}, under our approach the robot adapts to use the orange light to convey aggressive driving with one human, and the black light to convey the same data to another user.

Overall, we make the following contributions:

\p{Communication as Optimization} We formulate an optimization problem for identifying the interface signals in task-agnostic settings. Under this formulation the robot learns an interface mapping that (a) maximizes the correlation between the human's behavior and the information the robot is trying to convey, while also (b) conforming to user-friendly priors.

\p{Incorporating User-Friendly Priors} We propose intuitive priors that humans expect interfaces to follow (such as proportionality and convexity). The interface biases its signals towards these priors through a real-time learning approach that personalizes to the current human's behavior.

\p{Comparisons in Simulation and User Studies} 
We compare our resulting framework against state-of-the-art baselines and ablations of our own approach.
Across a $15$ person user study we find that participants prefer working with interfaces that leverage our approach, and that these interfaces learn to communicate the robot's information in ways that objectively improve the human's task performance.

\section{Related Work} \label{sec:related}

Prior research explores how robots can use interfaces to communicate information to human partners.
In settings such as robotic surgery
\cite{simorov2012review,rozeboom2014intuitive} or collaborative assembly
\cite{villani2018survey, cha2018survey}, these interfaces are typically pre-defined and held constant across all users.
Here \textit{the human must adapt to the interface} (i.e., the human must learn what meaning the robot ascribes to each signal).
Often these pre-defined interfaces are intuitive --- and require little human adaptation --- because they build upon socially accepted conventions.
For instance, in \cite{
weng2019robot, andersen2016projecting, walker2018communicating, mullen2021communicating} the robot communicates data to a human collaborator using signals that are unambiguous (e.g., an arrow pointing to the robot's goal).

However, when humans interact with learning systems or interfaces that integrate novel types of auditory, visual, and haptic feedback, it is not always obvious what the robot is attempting to convey \cite{belpaeme2018social,  gasteiger2021factors, dunkelberger2018conveying, valdivia2023wrapping}.
Within these contexts the \textit{interface should also adapt to the human}, and change the meaning of its signals to match the human's interpretations.
In what follows we survey two fundamental approaches for adapting and personalizing the robot's interface.

\p{Model-Based Interface Adaptation}
One group of related works automatically tunes the interface's signals based on either (a) models of the human user or (b) the human's task performance.
For example, in machine teaching a robot attempts to convey its latent state to a human: the robot has a model of how the human interprets its actions, and the robot reasons over this model to select communicative actions \cite{dragan2013legibility, huang2019enabling,  brown2019machine}.
When the human model changes --- e.g., the robot recognizes that the human is confused by complex signals --- the robot autonomously adapts its signals to align with the new model \cite{lee2021machine}.
In practice, however, the effectiveness of these machine teaching approaches is limited by the accuracy of their human models.
If a user deviates from the robot's expectations, the robot may unintentionally send signals that convey the incorrect information.
To address this challenge recent methods treat the human as a black box, and optimize the interface signals based on task performance \cite{kaupp2010human, reddy2022first}.
Here the robot trades-off between exploration and exploitation: the robot applies a set of signals, measures the human's achieved reward, and then decides whether to use those signals again or try different signal mappings.
But not only is this process potentially inefficient and time consuming, it can also lead to interfaces that over-fit to a given task.
Consider our driving example: if the interface tunes its mapping to help the human pass an autonomous car, that same interface may fail when the human is trying to merge behind the autonomous car.

\p{End-to-End Interface Adaptation}
A second body of research --- including our prior work \cite{christie2023limit} --- takes the opposite approach.
Instead of relying on a human model or given task, these methods use end-to-end structures to learn the interface mapping from scratch.
The advantage of these approaches is that the learned signals are task-invariant and highly customized to the current user.
But the downside is that end-to-end learning is data intensive: the robot requires many interactions with the current user to adapt its signals, and the resulting mappings may not be intuitive or user-friendly.
In this paper we seek to address this challenge by combining end-to-end learning with priors over the space of interface mappings.
Existing research in cognitive science suggests that humans have common biases in their communication
\cite{ramachandran2001synaesthesia,cwiek2022bouba},
and recent robotics research has applied similar priors to different problem settings \cite{li2020learning}.
Our hypothesis is that incorporating these priors will accelerate the robot's adaptation, leading to more effective communication interfaces than purely model-based or end-to-end approaches.

\section{Problem Statement}

We consider scenarios in which a robotic interface is trying to communicate hidden information to a human operator.
 For instance, in our running example an autonomous vehicle seeks to convey its driving style to a nearby human-driven car.
 The interface has access to the hidden information (e.g., the robot knows whether it is an aggressive or defensive driver), and the human needs this hidden information to perform their task correctly (e.g., the human must determine the robot's driving style to safely pass).
 Within this context, we specifically focus on settings where there are not obvious or mutually agreed upon signals for the robot to convey its hidden information.
 Our fundamental challenge is determining how --- in the absence of established conventions --- a robotic interface should adapt its signals over time to convey information to humans.
 
 \p{Repeated Interaction}
 We assume that the human works with the robotic interface multiple times.
 At the start of each interaction the robot observes the hidden information $\theta \in \Theta$ (e.g., parameters capturing the autonomous car's current driving style).
 This hidden information can change between interactions, but we assume that $\theta$ remains constant within an interaction.
 Each interaction lasts for a total of $T$ timesteps.
 
 \p{Interface} Let $s^t$ be the system state at timestep $t$. 
 Within our running example the system state is the position of both the autonomous car and the human-driven car.
 The robotic interface observes state $s$ and hidden information $\theta$, and then chooses signals $x \in \mathcal{X}$ according to its policy:
 \begin{equation}
     x \sim \pi_\mathcal{R}\left(\circ \mid s, \theta \right)
     \label{eq:interface-policy}
 \end{equation}
 where $x^t$ is the signal that the interface outputs at timestep $t$.
 Over the course of this interaction the interface displays these signals to the human, and the human uses these signals to infer $\theta$ and complete their task (e.g., safely driving around the autonomous car). We emphasize that this approach is collaborative: the interface seeks to render signals that are interpretable and meaningful to the user.
 
 \p{Human} The human operator cannot directly observe the hidden information $\theta$. 
 Instead, the human sees the interface signals $x$ and the system state $s$, and then interprets these signals to determine what actions they should take.
 Let $a^t$ be the human's action at timestep $t$, where this action is sampled from the human's policy (\eq{human-policy}). These actions cause the system state to transition (\eq{transition}):
 \begin{gather}
     a \sim \pi_\mathcal{H}\left(\circ \mid s, x\right)
     \label{eq:human-policy}
     \\
     s^{t+1} = f(s^t, a^t, u^t)
     \label{eq:transition}
 \end{gather}
where $u$ is the robot's action (e.g., the autonomous car accelerating or changing lanes).
In some of our experiments the robot does not take actions and $u = 0$.

Over repeated interactions, we seek to learn a \textit{task-agnostic} interface policy $\pi_\mathcal{R}$ that efficiently communicates $\theta$ to the human. The robot does not initially know how the human will interpret its signals, so the robot must \textit{adapt its signals} so that those signals are interpretable and meaningful for the current user.
 
 
\section{Using User-Friendly Priors to \\ Accelerate Signal Adaptation}\label{sec:method}

The interface is not sure how the human will interpret its signals.
Here we apply our hypothesis that --- even though different humans interpret the same signals in different ways --- there exist underlying patterns that all users expect signals to follow.
Returning to our example where the autonomous car displays its driving style with an LED light: we do not know if the human will interpret an orange light as an aggressive or defensive driver --- but we can anticipate that the human will interpret these signals according to previous expectations on interface design.
We encode this expectation as a \textit{prior} over the space of interface policies $\pi_\mathcal{R}$.
By itself, this prior is still not sufficient for the interface to identify the correct $\pi_\mathcal{R}$ for the current user.
However, we will leverage this prior to narrow down the range of mappings the interface explores and ultimately accelerate the co-adaptation between the learning interface and human operator.

In Section~\ref{sec:m2} we theoretically derive how priors can be incorporated in the interface-adaptation algorithm \textbf{LIMIT} \cite{christie2023limit} to accelerate human-robot co-adaptation, then in Section~\ref{sec:m3} we forumate intuitive priors that humans might expect their interfaces to follow (e.g., proportional and convex signals).

\subsection{Learning Interface Policies that Incorporate Priors} 
\label{sec:m2}

The interface-adaptation algorithm \textbf{LIMIT} proposed in \cite{christie2023limit} attempts to maximize the correlation between the user's state-actions and the hidden information that the interface is trying to convey.
To summarize, \textbf{LIMIT} maximizes \textit{decodability}: a proxy to conditional mutual information (\eq{cmi-def}).

\begin{equation}
    I(a \,;\, \theta \mid s) = H(a \mid s) - H(a \mid \theta, s)
\label{eq:cmi-def}
\end{equation}
Maximizing correlation offers a first-pass solution for learning an interpretable interface policy.
However, mutual information alone does not take advantage of our hypothesis that there are user-friendly priors operators expect their interfaces to follow.
For example, if we optimize for \eq{cmi-def} alone the autonomous car could learn to use an orange LED light to convey aggressive driving, a partially illumined LED light for defensive driving, and then turn the light off for balanced aggressive-defensive driving. 
Although users might adapt to this interface given enough time, it is not intuitive: it would better align with our expectations for the extremes of the signal to capture the extremes of the robot's behavior.

We here encode these types of expectations as a prior $P_0(x \mid s, \theta)$.
This prior captures how the human might expect the robotic interface to map hidden information to signals $x$.
Following  \cite{cover2012elements}, we note that the correlation between $a$ and $\theta$ with the prior $P_0$ is \textit{greater than or equal to} the correlation between $a$ and $\theta$ without this prior, given that the robotic interface and the human observe the prior $P_0$:
\begin{equation} \label{eq:m1}
I(a \,;\, \theta \mid s, P_0) \geq I(a \,;\, \theta \mid s)
\end{equation}
Comparing \eq{m1} to \eq{cmi-def}, we see that incorporating priors theoretically increases the correlation between human actions and hidden information.
To now reformulate $I(a \,;\, \theta \mid s, P_0)$ as an optimization problem over the space of interface policies, we introduce the objective:
\begin{equation}
    \argmax_{\pi_\mathcal{R}}  \Big[I(a \,;\, \theta \mid s) - D_{KL}(\pi_\mathcal{R} \| P_0 )\Big]
    \label{eq:opt}
\end{equation}
where $D_{KL}(\pi_\mathcal{R} \| P_0 )$ is the Kullback-Leibler divergence between the interface distribution $\pi_\mathcal{R}$ and the prior distribution $P_0$.
Re-writing this optimization problem in terms of the interface and human policies, we reach:
\begin{equation} \label{eq:m2}
    \argmax_{\pi_\mathcal{R}} ~ \int\limits_{x, s, \theta}
    \Big(
\int\limits_{a}
\Big( p(a, s, x, \theta) \log T \Big) + \pi_\mathcal{R} \log P_0 \Big)
\end{equation}
In the above we abbreviate $\pi_\mathcal{R}(x \mid s, \theta)$ as $\pi_\mathcal{R}$ and $\pi_\mathcal{H}(a \mid s, x)$ as $\pi_\mathcal{H}$. The term $T$ is defined as: 
\begin{equation}
\label{eq:m3}
    T = 
\int_{x^\prime} \pi_\mathcal{H} \pi_\mathcal{R}
\Bigg(\int_{x^\prime} 
\pi_\mathcal{H} \pi_\mathcal{R}
\int_{\theta^\prime} \pi_\mathcal{R} p(\theta^\prime) \Bigg)^{-1}
\end{equation}
Equations~(\ref{eq:m2}) and~(\ref{eq:m3}) capture the objectives of \eq{opt}: (a) maximizing correlation between human and interface while (b) biasing the interface towards the prior.
To tractably solve Equations~(\ref{eq:m2}) and (\ref{eq:m3}) we return to the neural networks $\mathcal{R}_\psi$, $\mathcal{H}_\phi$, and $\Delta_\sigma$ from \cite{christie2023limit}.
Let $\mathcal{D} = \{(s^1, a^1, x^1, \theta^1), \ldots , (s^n, a^n, x^n, \theta^n)\}$ be a dataset of $n$ past tuples.
The interface continually collects these tuples over repeated interactions as it collaborates with the current user.
We will leverage the dataset $\mathcal{D}$ to adjust our models in a way that approximately maximizes $I(a \,;\, \theta \mid s, P_0)$.

\p{Correlation Loss}
Noting that the term $T$ in \eq{m3} is equivalent to the conditional mutual information proxy originally proposed in \cite{christie2023limit}, we will use the \textbf{LIMIT} loss functions $\mathcal{L}_\pi \left(\phi\right)$ and $\mathcal{L}_\Delta\left(\psi, \sigma\right)$ to update the policy network weights $\phi$ and $\psi$ and the decoder weights $\sigma$:
\begin{equation}
\mathcal{L}_\pi (\phi)
=
\sum_{(s, a, \theta) \in \mathcal{D}}
\Big\| a - 
\mathcal{H}_\phi\left(s, \mathcal{R}_\psi\left(s, \theta\right)\right)
\Big\|^2
    \label{eq:human-loss}
\end{equation}
\eq{human-loss} asserts that the human model should map the interface's signal to the actual human behavior $a$;
\begin{equation}
\begin{gathered}
\tau\left(s, \theta\right) = \{(s, a)^0, \ldots, (s, a)^k\} 
\\
a^t = \mathcal{H}_\phi\left(s^t, \mathcal{R}_\psi\left(s^t, \theta\right)\right), \quad s^{t+1} = f\left(s^t, a^t\right)
\\
    \mathcal{L}_\Delta\left(\psi, \sigma\right)
    =
    \sum\limits_{(s, \theta) \in \mathcal{D}}
    \Big\|
    \theta - \Delta_\sigma\left(\tau\left(s, \theta\right)\right)
    \Big\|^2
\end{gathered}
    \label{eq:interface-loss}
\end{equation}
whereas \eq{interface-loss} attempts to minimize the denominator of $T$: maximizing the likelihood that humans take unique actions for specific hidden information.
\p{Prior Loss} To approximate the $\pi_\mathcal{R} \log P_0$ term that appears in \eq{m2}, we introduce the prior loss $\mathcal{L}_P$:
\begin{equation}
    \label{eq:L1}
    \mathcal{L}_P(\psi) = \sum_\mathcal{D} ~ \|  \hat{x} - \mathcal{R}_\psi(s, \theta) \|^2, \quad \hat{x} \sim P_0(\circ \mid s, \theta)
\end{equation}
Minimizing \eq{L1} modifies the interface policy to match the prior; this is analogous to maximizing $\pi_\mathcal{R} \log P_0$.

\p{Overall Loss} Putting these three loss functions together to approximate Equations~(\ref{eq:m2}) and (\ref{eq:m3}), we reach the overall loss function $\mathcal{L}$ used to train our models:
\begin{equation}
\mathcal{L}\left(\psi, \phi, \sigma\right)
= 
\lambda_1 \mathcal{L}_P\left(\psi\right) + 
\lambda_2 \mathcal{L}_\pi \left(\phi\right) +
\lambda_3 \mathcal{L}_\Delta \left(\psi, \sigma\right)
\label{eq:all-loss}
\end{equation}
where $\lambda_1$, $\lambda_2$, and $\lambda_3$ are design parameters in $[0, \infty)$. 
In practice, the robot updates $\mathcal{R}_\psi$, $\mathcal{H}_\phi$, and $\Delta_\sigma$ online in order to minimize $\mathcal{L}$.
The interface then applies the learned policy $\mathcal{R}_\psi$ to choose the signals that it conveys to the human.

\begin{figure*}[t]
    \centering
    \includegraphics[width=1.8\columnwidth]{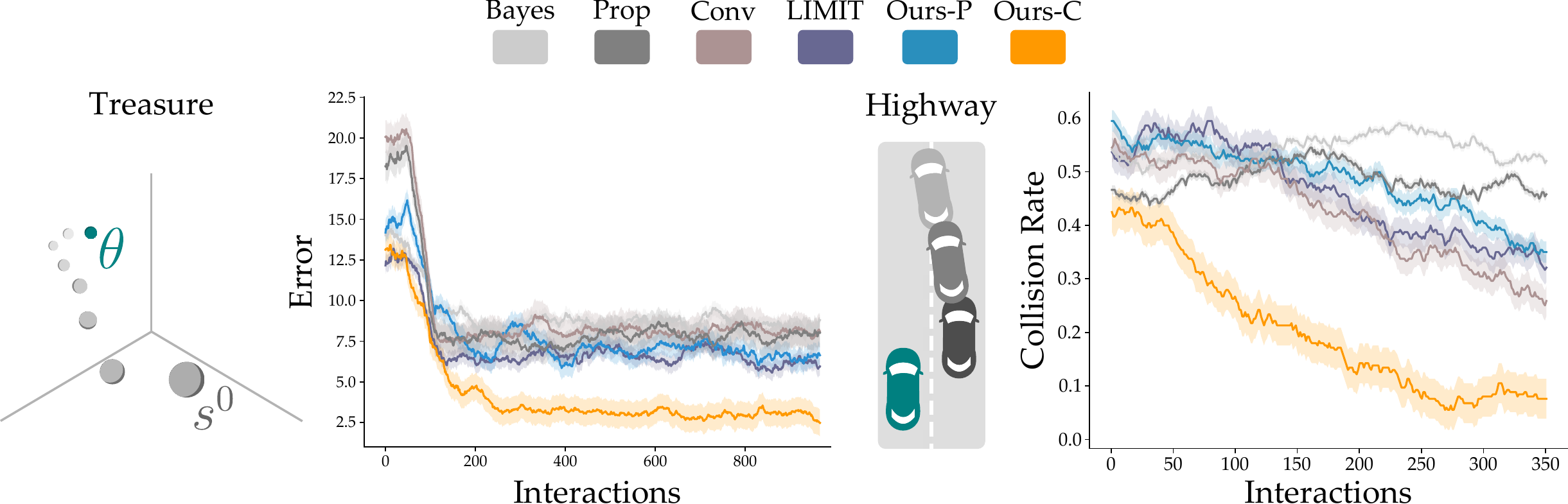}
    \caption{
    Results from the \textbf{Treasure} (left) and \textbf{Highway} (right) simulations.
    \textbf{Treasure}: The human starts in state $s^{0}$ and tries to reach a hidden state $\theta$ that only the robot knows. The results shown here are averaged across $5$ simulations of $1000$ interactions in a $3$D environment. These results are consistent from $2$D to $8$D environments. Ours-C outperforms all other baselines in terms of average error ($p < 0.001$).
    \textbf{Highway}: The human (teal) tries to pass the autonomous vehicle (grey) without a collision. Here the results are averaged over $10$ simulations of $350$ interactions. Ours-C significantly outperforms baselines and Ours-P ($p < 0.001$). 
    We note that the combination of \textit{Proportionality} and \textit{Convexity} are not shown: we found that these two priors conflict with one another during learning. Intuitively, this occurs because \textit{Convexity} attempts to spread signals out, while \textit{Proportionality} groups signals together. 
    }
    \label{fig:sims}
    \vspace{-1.5em}
\end{figure*}

\subsection{User-Friendly Interface Priors}\label{sec:m3}

In Section~\ref{sec:m2} we presented an information theoretic approach to learn interface policies that are interpretable for the current user and biased towards a given prior.
However, it is still not clear what these priors should be. 
Put another way, what distribution(s) should the interface use for $P_0$ in \eq{L1}?
Below we propose two user-friendly properties that can be leveraged to derive interface priors.
Our approach is not tied to these specific priors --- the method that we have developed can be applied to any prior distribution provided by the designer.
However, our experiments will focus on the two priors described below: proportionality and convexity.

\p{Proportionality} 
Intuitively, changes in signals $x$ should be \textit{proportional} to changes in hidden information $\theta$. We enforce this intuition by choosing $\mathcal{L}_P\left(\psi\right)$ as:
\begin{equation}
    \sum_{(s, \theta_1, \theta_2) \in \mathcal{D}}
    e^{\gamma \|\theta_1 - \theta_2 \|^2}
    \Big\|
    \mathcal{R}_\psi\left(s, \theta_1\right) - \mathcal{R}_\psi\left(s, \theta_2\right)
    \Big\|^2
\label{eq:prop-loss}
\end{equation}
where $\gamma$ is a hyperparameter that tunes the sensitivity to changes in $\theta$.
This choice of $\mathcal{L}_P$ scales the difference in signals proportionally to the difference in hidden information.

\p{Convexity} 
Returning to our running example, at one extreme of $\theta$ the autonomous car is fully aggressive, and at the other extreme of $\theta$ the autonomous car is fully defensive.
Our intuition is that these opposite values of $\theta$ should correspond to opposite signals $x$; e.g. if the LED light is fully illuminated to signal an aggressive car, then it is turned off to signal the defensive car.
We formalize this user-friendly prior by making the space of signals \textit{convex} in magnitude.
To make $\| \mathcal{R}_\psi\left(s, \theta\right)\|$ a convex function of $\theta$, we select $\mathcal{L}_P\left(\psi\right)$ to be:
\begin{equation}
    \mathcal{L}_P(\psi) = \sum\limits_{(s, \theta) \in \mathcal{D}}
    \Big\|
    \mathcal{R}_\psi\left(s, \theta\right) + \mathcal{R}_\psi\left(s, -\theta\right)
    \Big\|^2
    \label{eq:convex-set-loss}
\end{equation}
This choice of $\mathcal{L}_P$ together with \eq{interface-loss} ensures that $\|\mathcal{R}_\psi\left(s,\theta\right)\|$ forms a convex set with a minimum at $\theta = 0$. 

\p{Summary}
To summarize our proposed approach, we use the combined loss function in \eq{all-loss} to tractably approximate \eq{opt}. 
Optimizing this loss function produces interface policies that (a) maximize the correlation between human actions and $\theta$, and (b) minimize the divergence from human-friendly priors (e.g., proportionality or convexity).
In practice, the interface is initialized with a policy consistent with the prior and then starts collaborating with the human.
As the interface gathers data from the human operator, it updates the dataset $\mathcal{D}$ and retrains the interface policy online to minimize the loss $\mathcal{L}$ across $\mathcal{D}$.
This leads to an interface that adapts its signals over time to establish correlation with the current user, while also maintaining the intuitive properties that humans expect interfaces to follow.

\section{Simulations}\label{sec:sims}


We first compare our proposed approach for adapting interface signals to state-of-the-art baselines \cite{bayes, christie2023limit} and ablations of our method across controlled simulations. We consider two different environments --- Treasure and Highway --- described later in this section. In each environment, the interface knows the hidden information $\theta$ (e.g., location of the treasure) and must generate signals to communicate this hidden information to a simulated human. The simulated humans are adaptive agents that change how they interpret the signals between interactions. Accordingly, the interface must adapt its signals to align with the simulated human in order to accurately convey $\theta$.
Code for implementing these simulations can be found here: {\urlstyle{same} \url{https://github.com/VT-Collab/interfaces-human-priors}}.

\p{Interface Algorithms} We compare the following methods for adapting interface signals:
\begin{itemize}
    \item \textbf{Bayes}: The interface generates signals according to $x = \bm{A} \begin{bmatrix} s & \theta \end{bmatrix}^T$, where $\bm{A}$ is a matrix found using \textit{Bayesian Optimization} \cite{reddy2022first, bayes}. The interface adjusts $\bm{A}$ to maximize the total observed reward.
    \item \textbf{Prop}: The interface generates signals \textit{only} according to the \textit{Proportionality} prior from Section \ref{sec:m3}
    \item \textbf{Conv}: The interface generates signals \textit{only} according to the \textit{Convexity} prior from Section \ref{sec:m3}
    \item \textbf{LIMIT}: 
    The interface generates signals to maximize the correlation between human and robot using an existing end-to-end approach~\cite{christie2023limit}.
    \item \textbf{Ours-P}: The interface generates signals according to our proposed approach with the \textit{Proportionality} prior.
    \item \textbf{Ours-C}: The interface generates signals according to our proposed approach with the \textit{Convexity} prior.
\end{itemize}
Note that --- unlike the other methods --- \textbf{Bayes} has access to the human's reward function and uses it to optimize the robot's interface.
This gives \textbf{Bayes} more domain knowledge than the alternative approaches.

\p{Simulated Human} We pair each algorithm with a simulated agent. 
These simulated agents are multi-layer perceptrons consistent with \eq{human-policy}:  $\mathcal{M}: \mathcal{X} \times \mathcal{S} \mapsto \mathcal{A}$.
To enforce human-like biases in signal interpretation, we pretrain the human with corresponding interface structures (\textbf{Bayes}, \textit{Convexity}, etc.).
This is separate from the testing regime; simulated humans are not tested with pretrained interfaces. 


\p{Treasure} In this environment the simulated human is navigating a continuous space in $\mathbb{R}^n$, and is attempting to reach the treasure at $s = \theta$ within $10$ timesteps. The human state $s$ and treasure location $\theta$ are randomly sampled from $U([-10, 10]^{n})$ at the start of each interaction. The human cannot observe $\theta$ and must rely on the $n$-dimensional interface signals to infer the hidden location and choose goal-directed actions.
We test the interface algorithms in environments ranging from two to eight dimensions (shown with $n=3$ in the left side of Figure~\ref{fig:sims}). We measure performance as $\|s^T - \theta\|^2$. The results indicate that \textbf{Ours-C} better communicates the hidden goal location than the alternatives ($p < 0.001$).

\p{Highway} In this environment the simulated human is driving along a two-lane one-way highway. In front of the human is an autonomous vehicle that the human must avoid. The autonomous vehicle's policy is parameterized by $\theta$ and chosen from several discrete policies: 

\begin{enumerate}[label=(\alph*)]
    \item The autonomous car will always stay in the right lane
    \item The autonomous car will always stay in the left lane
    \item The autonomous car will merge into the lane of the human at the previous timestep
    \item The autonomous car will merge into the \textit{opposite} lane of the human at the previous timestep
\end{enumerate}
The autonomous car's driving style (e.g., policy parameters $\theta$) is sampled before the start of an interaction. Every interaction lasts for $10$ timesteps. At each time step, the autonomous vehicle attempts to signal its policy using a $1$D interface with signals $x \in \mathbb{R}$. 
The human must infer the next action of the autonomous vehicle from its signals and choose an action that avoids collision. 
Figure \ref{fig:sims} shows that \textbf{Ours-C} outperforms other interface algorithms, achieving a lower mean collision rate $\sum\left(\mathbbm{1}[s_\mathcal{R}^t = s_\mathcal{H}^t]\right) / T$ ($p < 0.001$).
\section{User Study} \label{sec:user-study}

\begin{figure*}[t!]
    \centering
    \includegraphics[width=1.8\columnwidth]{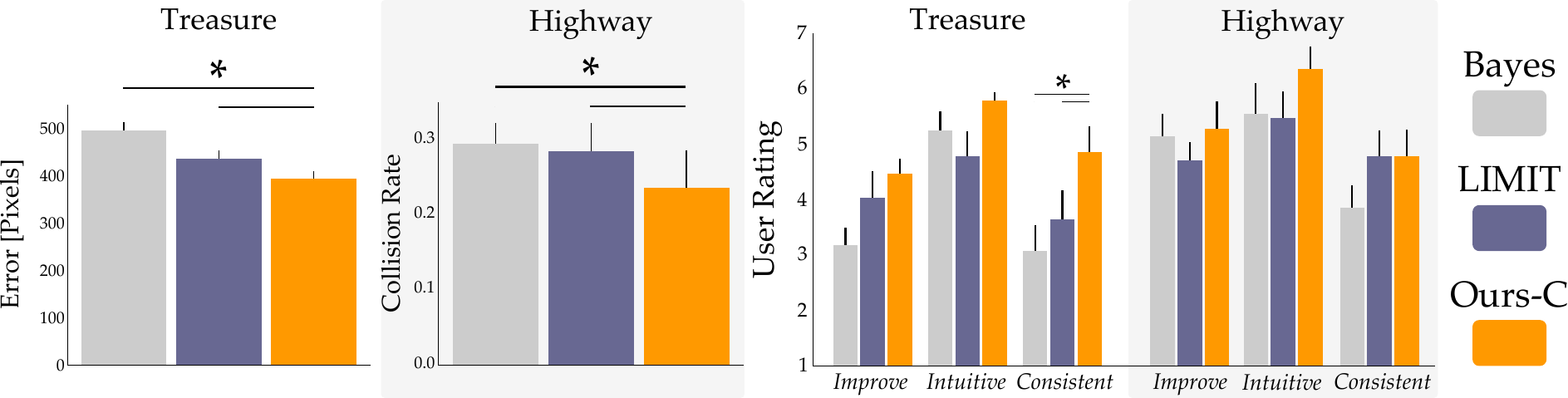}
    \caption{Left: Ours-C enables users to achieve a better performance in the Treasure (left) and Highway (right) tasks than LIMIT or Bayes. The difference in performance between methods is significant ($p < 0.0001$ (Treasure), $p \to 0.05$ (Highway)). An asterisk (*) denotes significance.
    Right: Users indicated that interfaces generated using the convexity prior (Ours-C) are more \textit{consistent} and \textit{intuitive}, and that they \textit{improved} more over time with Ours-C.}
    \label{fig:user-study}
    \vspace{-1.5em}
\end{figure*}

Our controlled simulations suggest that incorporating a convexity prior over the space of interface mappings accelerates adaptation. 
To evaluate how Ours-C performs with real users, we next conducted an in-person study with a robotic interface.
This study took place using the same environments from Section~\ref{sec:sims}.
In each task the interface knew the hidden information $\theta$ (e.g., the policy of an autonomous vehicle), and selected signals to convey $\theta$ to the actual human. 

\p{Independent Variables} We compared three algorithms for generating signals: \textbf{Bayes}, \textbf{LIMIT}, and \textbf{Ours-C}. As described in Section \ref{sec:sims}, \textbf{Bayes} observes the human's reward at the end of each interaction and updates its mapping from $(s, \theta)$ to signals to maximize this reward. \textbf{LIMIT} learns signals end-to-end to maximize the correlation between $\theta$ and the human actions. Finally, \textbf{Ours-C} uses our proposed approach to optimize for correlation while also minimizing deviation from the convexity prior in Section \ref{sec:m3}. 

\p{Experimental Setup} We divided the study into two tasks: \textbf{Treasure} and \textbf{Highway}. These tasks were identical to their simulated counterparts. A collection of monochromatic bars of varying color were used to communicate the interface's signals to the human, similar to the interface shown in Figure \ref{fig:front}. This signal was one-dimensional for the Highway task and two-dimensional for the Treasure task.

%

\p{Participants and Procedure}
We recruited 15 participants (2 female, age $25.9 \pm 5.2$ years) from the Virginia Tech community. All participants provided informed written consent as per the university guidelines (IRB \#20-755). 

Each participant completed the tasks three times: once for each method tested. Both the order of methods tested and the order of tasks completed were counterbalanced. Participants were never told which algorithm they were working with.

\p{Dependent Measures --- Objective} We recorded the states, signals, actions, and hidden information during each interaction. 
To assess user performance in the \textbf{Treasure} and \textbf{Highway} tasks, we calculate final state error and the collision rate of each interaction respectively. Lower values for final state error and collision rate indicate that the user correctly inferred the hidden information from the interface's signals.

\p{Dependent Measures --- Subjective} After each task and algorithm, participants completed a 7-point Likert scale survey. This survey measured the user's subjective preferences along three multi-item scales. We asked users:
\begin{enumerate}
    \item If they felt like their performance \textit{improved} over time,
    \item If they thought that the interface's signals were \textit{intuitive},
    \item If the interface's signals were \textit{consistent}.
\end{enumerate}
Altogether, these questions gauged how well users thought they collaborated with the interface. 

\p{Hypotheses} We had two hypotheses for the user study:
\begin{quote}
\p{H1} \textit{
Users will perform the task better when receiving signals from an interface that adapts with
} 
\textbf{Ours-C} 
\textit{than with either}
\textbf{Bayes} \textit{or} \textbf{LIMIT}\textit{.}
\end{quote}
\begin{quote}
\p{H2} \textit{Users will subjectively prefer interfaces that adapt using} \textbf{Ours-C}\textit{.}
\end{quote}

\p{Results} The objective results of our user study are summarized in Figure \ref{fig:user-study} (left).
To address \textbf{H1}, we measured the final state error for \textbf{Treasure} and the collision rate for \textbf{Highway}. For both metrics, lower values indicate better performance. An effective interface will help the human in navigating to the position of the treasure or minimizing their collision with the autonomous car. One-way ANOVA tests showed that participants had significantly lower error when working with interfaces generated by \textbf{Ours-C} in both the Treasure ($F(3, 627)=9.361, p < 0.001$) and Highway ($F(3,837)=2.971, p\to0.05$) environments. These results indicate that interfaces generated with \textbf{Ours-C} provided more helpful signals than the \textbf{Bayes} and \textbf{LIMIT} baselines. 

Regarding \textbf{H2}, we present the subjective results from our Likert-scale survey in Figure \ref{fig:user-study} (right). 
A one-way ANOVA analysis of the users' responses showed a significant difference in the perceived \textit{Consistency} of the interface signals in the Treasure task ($F(2, 39) = 3.919, p < 0.05$). 
Although we did not see a significant difference in user ratings for the other items, overall, users stated that they preferred interfaces generated with \textbf{Ours-C} in both tasks.
Compared to the baselines, the signals generated by \textbf{Ours-C} were more consistent with user expectations  --- generating opposite signals for opposite values of $\theta$ by optimizing \eq{convex-set-loss}.

\section{Conclusion}

In this paper we studied settings where a robotic interface is attempting to convey information to a human, and there are no pre-existing conventions for how the human should interpret the interface's signals.
Although different humans may interpret the same signal in different ways, we hypothesized that all humans have underlying patterns they expect the interface to follow. 
We leveraged this insight to propose an information theoretic approach that maximizes the correlation between interface and human while biasing the interface towards user-friendly priors.
In practice, this resulted in self-adjusting interfaces that adapted their own signals to better convey information to the human.
When compared to model-based and end-to-end baselines across simulations and a user study with $15$ participants, we found that incorporating priors accelerated adaptation and improved communication from robot to human.


\balance
\bibliographystyle{IEEEtran}
\bibliography{IEEEabrv,bibtex}

\end{document}